\documentclass[sigconf]{acmart}
\usepackage{graphicx}
\usepackage[caption=false]{subfig}
\usepackage[]{algorithm2e}

\acmYear{}
\acmPrice{}
\acmISBN{}
\acmDOI{}
\setcopyright{none}
\acmConference[KDD-Feed'19]{Feed'19: Workshop on Intelligent Information Feed}{August 5th, 2019}{Anchorage, Alaska, USA}

\begin{document}

\title[SEW: Generating Glyph for Sentiment Analysis]{Squared English Word: A Method of Generating Glyph to Use Super Characters for Sentiment Analysis}

\author{Baohua Sun}
\email{baohua.sun@gyrfalcontech.com}
\affiliation{%
  \institution{Gyrfalcon Technology Inc.}
  \streetaddress{1900 McCarthy Blvd.}
  \city{Milpitas}
  \state{California}
  \postcode{95035}
}

\author{Lin Yang}
\affiliation{%
  \institution{Gyrfalcon Technology Inc.}
  \streetaddress{1900 McCarthy Blvd.}
  \city{Milpitas}
  \state{California}
  \postcode{95035}
}

\author{Catherine Chi}
\affiliation{%
  \institution{Gyrfalcon Technology Inc.}
  \streetaddress{1900 McCarthy Blvd.}
  \city{Milpitas}
  \state{California}
  \postcode{95035}
}

\author{Wenhan Zhang}
\affiliation{%
  \institution{Gyrfalcon Technology Inc.}
  \streetaddress{1900 McCarthy Blvd.}
  \city{Milpitas}
  \state{California}
  \postcode{95035}
}

\author{Michael Lin}
\affiliation{%
  \institution{Gyrfalcon Technology Inc.}
  \streetaddress{1900 McCarthy Blvd.}
  \city{Milpitas}
  \state{California}
  \postcode{95035}
}



\renewcommand{\shortauthors}{Sun and Yang, et al.}

\begin{abstract}
The Super Characters method addresses sentiment analysis problems by first converting the input text into images and then applying 2D-CNN models to classify the sentiment. It achieves state of the art performance on many benchmark datasets. However, it is not as straightforward to apply in Latin languages as in Asian languages. Because the 2D-CNN model is designed to recognize two-dimensional images, it is better if the inputs are in the form of glyphs. In this paper, we propose SEW (Squared English Word) method generating a squared glyph for each English word by drawing Super Characters images of each English word at the alphabet level, combining the squared glyph together into a whole Super Characters image at the sentence level, and then applying the CNN model to classify the sentiment within the sentence. We applied the SEW method to Wikipedia dataset and obtained a 2.1\% accuracy gain compared to the original Super Characters method. For multi-modal data with both structured tabular data and unstructured natural language text, the modified SEW method integrates the data into a single image and classifies sentiment with one unified CNN model.
\end{abstract}





\maketitle
\section{Introduction}

The need to classify sentiment arises in many different problems in customer related marketing fields. Super Characters \cite{sun2018super} is a two-step method for sentiment analysis. It first converts text into images; then feeds the images into CNN models to classify the sentiment. Sentiment classification performance on large text contents from customer online comments shows that the Super Character method is superior to other existing methods, including fastText\cite{joulin2016bag}, EmbedNet, OnehotNet, and linear models\cite{zhang2017encoding}. 

\begin{figure}
\centering
\subfloat[Raw Super Characters method for English sentence on alphabet level.\label{EnglishTextDemonstration_a}]{
\includegraphics[width=.20\textwidth]{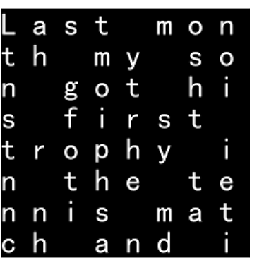}
}~~~~~~~~~~~~~~~~
\subfloat[Raw Super Characters method for English sentence on alphabet level with changing line to avoid breaking words. \label{EnglishTextDemonstration_b}]{
\includegraphics[width=.20\textwidth]{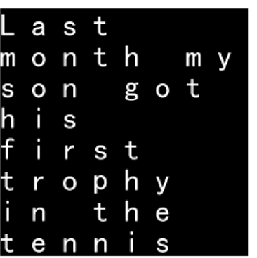}
}\\
\subfloat[Squared English Word (SEW) method with 6x6 words per image.\label{EnglishTextDemonstration_c}]{
\includegraphics[width=.20\textwidth]{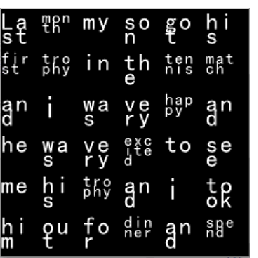}
}~~~~~~~~~~~~~~~~
\subfloat[Squared English Word method with attention on the first four words. \label{EnglishTextDemonstration_d}]{
\includegraphics[width=.20\textwidth]{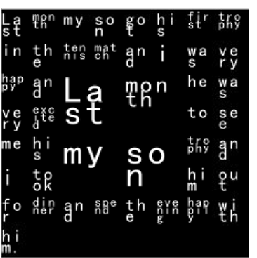}
}\\
\subfloat[Squared English Word method using both happy moment text and profiles information, age 36, country India (IND), married (m), male (m). \label{EnglishTextDemonstration_f}]{
\includegraphics[width=.20\textwidth]{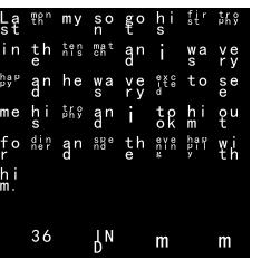}
}~~~~~~~~~~~~~~~~
\subfloat[Squared English Word method using happy moment text and attended profile, age 36, country India (IND), married (m), male (m).\label{EnglishTextDemonstration_e}]{
\includegraphics[width=.20\textwidth]{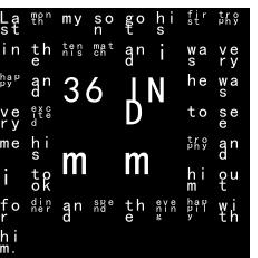}
}
\caption{Demonstrations of raw Super Characters method and Squared English Word method. We use the same example input to illustrate our idea. The raw text input sentence is: ``Last month my son got his first trophy in the tennis match and i was very happy and he was very excited to see me his trophy and i took him out for dinner and spend the evening happily with him."\label{EnglishTextDemonstration}}
\end{figure}

However, there are a few challenges of using the Super Characters method for Latin language inputs. First, the Super Characters method can be directly applied for Asian languages with glyph characters, such as Chinese, Japanese, and Korean, but not so in such a straightforward fashion to Latin languages such as English. This is because the CNN model connected to the super characters images are designed to recognize two-dimensional images better in the form of a glyph in a square form. Languages like Chinese build their language system upon logograms, which are symbols or characters that serve to represent a phrase or word. If we directly apply Super Characters method to represent sentences in the English language, the Super Characters image is shown as in Figure \ref{EnglishTextDemonstration_a}. Or, as shown in Figure \ref{EnglishTextDemonstration_b} if we try to avoid breaking the words and changing lines because a word is divided between two lines, it will become harder for the CNN model to recognize.  
In Figure \ref{EnglishTextDemonstration_c}, we convert each English word to a glyph, such that each word only occupies the pixels within a designated squared area. The resulting algorithm is named Squared English Word (SEW). In the original Super Characters method, all the text drawn on the image are given the same size, or given the same degree of attention when it is fed into the CNN model connected to it. We add the attention scheme by allocating larger spaces for important words, e.g. those in beginning of each sentence. SEW with attention as shown in Figure \ref{EnglishTextDemonstration_d}. 

The CL-AFF Shared Task\cite{CL-AFF} is part of the Affective Content Analysis workshop at AAAI 2019. It builds upon the HappyDB dataset\cite{asai2018happydb_new}, which contains 10,560 samples of happy moments. Each sample is a text sentence describing the happy moments in English. And each sample has two sets of binary classification labels, Agency?(Yes$|$No) and Social?(Yes$|$No). In this paper, we will apply SEW and SEW with attention on this data set to classify the input texts.

\section{Squared English Word method}
The original Super Characters method works well if the character in that language is a glyph, and Asian characters in Chinese, Japanese, and Korean are written in a square form. In this work, we extend the original idea of Super Characters \cite{sun2018super} by preprocessing each English word into a squared glyph, just like Asian characters. To avoid information loss, the preprocessing should be a one-to-one mapping, i.e. each original English word can be recovered from the converted squared glyph. For text classification task, we propose the SEW method for English sentence input as described in Algorithm \ref{SquareEnglishWordAlgorithm}.

\begin{algorithm}[h]
 \KwIn{text input: a string of English words}
 \KwOut{Sentiment Classification Result}
 Initialization: start a blank image and set the font to draw Super Characters with, set a cut-length of the words, set counter=0, set current\_word=the first word in the text input, set the current\_word\_location for the current\_word which is a square area, and get the current\_word\_area as the area of pixels for current\_word\_location\;
 \While{not at end of the input text and counter$<$cut-length}{
  get the current\_word, set current\_alphabet=the first alphabet in the current\_word\;
  get current\_word\_length, set location\_stepsize=sqrt(alphabet\_area) where alphabet\_area is current\_word\_area divided by current\_word\_length, and set the current\_alphabet\_location for current\_alphabet at the top-left of the squared area of the current\_word\;
  \While{not at end of the current\_word}{
   draw the current\_alphabet at current\_alphabet\_location\;
   move to the next alphabet\ and update current\_alphabet\;
   update current\_alphabet\_location\ by moving one location\_stepsize, or change line if necessary\;
    
  }
   move to the next word\;
   counter+=1\;
 }
 Feed into CNN models, such as ResNet-50, and etc.\;
  \,
\Return{Sentiment Classification Result}\;
 \caption{Super Characters with SEW}\label{SquareEnglishWordAlgorithm}
\end{algorithm}

The proposed SEW method has shown accuracy improvement on DBpedia dataset provided in \cite{zhang2015character}, as shown in Table \ref{Table_1}.
DBpedia is a text classification dataset crawled from Wikipedia. It has 14 ontologies, each having 40,000 labeled text in training and 5,000 in testing.

\begin{table}[h!]
\begin{center}
\begin{tabular}{|c|c|}
\hline \bf Model & \bf Accuracy \\ \hline
SC\cite{sun2018super} & 96.2\% \\ \hline
SEW (this paper) & \bf 98.3\% \\
\hline
\end{tabular}
\end{center}
\caption{\label{Table_1} Results of our Squared English Word (SEW) method against original Super Characters (SC) method on DBpedia \cite{zhang2015character} data set. It shows the effectiveness of the squared method which improves performance by 2.1\%. The cut-length is set at 14x14=196 in order for the input with different length to fit. The CNN model used is SE-net-154\cite{SENet154}.}
\end{table}
\begin{figure}
\centering
\subfloat[Street Sign for ``East Main Street".\label{Fig_Street_a}]{
\includegraphics[width=.20\textwidth]{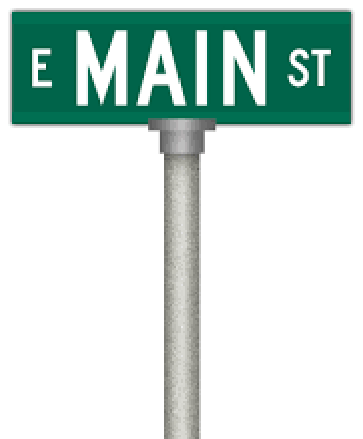}
}~~~~~~~~~~~~~~~~
\subfloat[Street Sign for ``Speed Limit 55" mph. \label{Fig_Street_b}]{
\includegraphics[width=.20\textwidth]{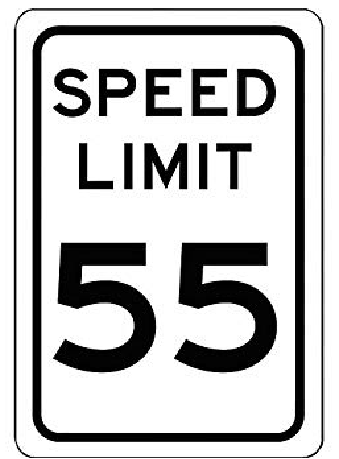}
}
\caption{Street sign example: the enlarged portions of the signs get attention\cite{StreetSign55MPH,StreetSignMainStreet}.}
\label{Fig_Street}
\end{figure}
\begin{figure*}
\centering
\subfloat[Histogram of CL-AFF Train Dataset.\label{Fig_TrainData_histogram}]{
\includegraphics[width=.36\textwidth]{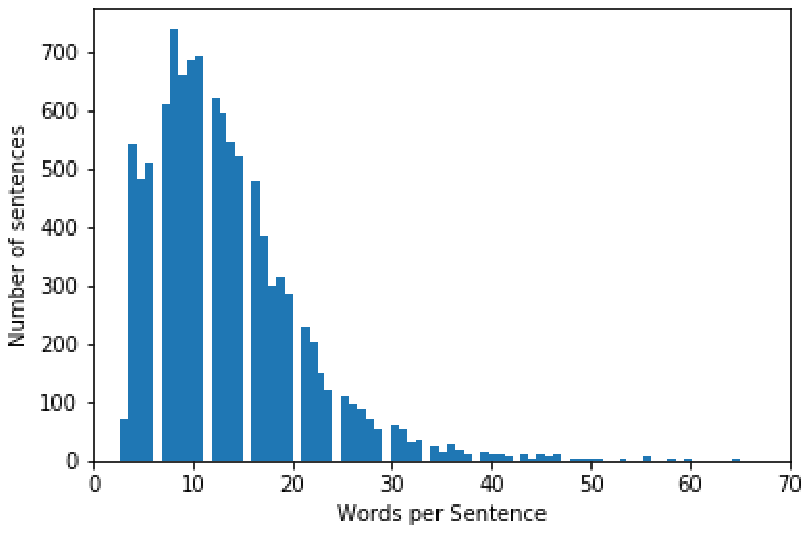}
}~~~~
\subfloat[Histogram of CL-AFF Test Dataset. \label{Fig_TestData_histogram}]{
\includegraphics[width=.36\textwidth]{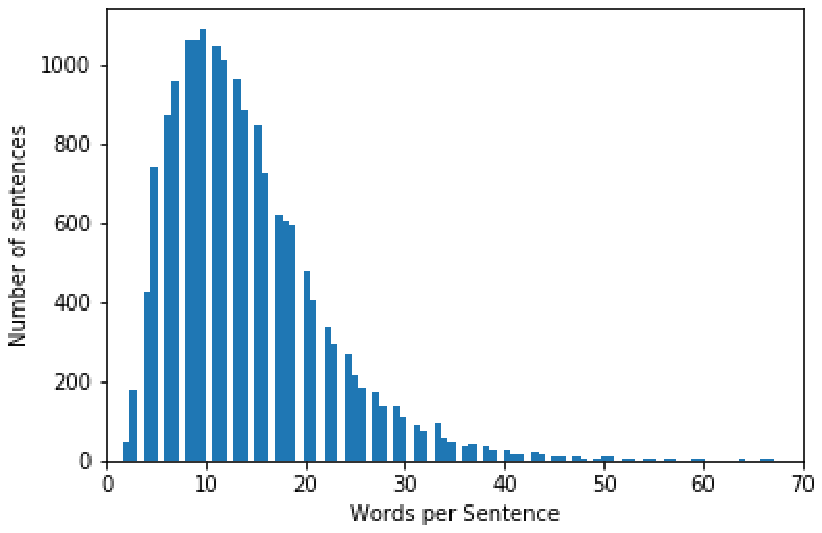}
}
\caption{Histogram of CL-AFF Dataset.}
\end{figure*}

\begin{table*}[h!]
\begin{center}
\begin{tabular}{|c|c|c|c|c|c|c|}
\hline  ~ & \bf median & \bf median & \bf max & \bf percentile of 25 & \bf percentile of 64 \\ \hline
Train & 13.44 & 12 & 70 & 92.76\% & 99.95\% \\ \hline
Test & 14.28 & 13 & 138 & 90.86\% & 99.84\% \\
\hline
\end{tabular}
\end{center}
\caption{\label{Table_stats} The statistics of training and testing data set.}
\end{table*}
\begin{table*}[h!]
\begin{center}
\begin{tabular}{|c|c|c|}
\hline \bf Approaches &\bf Agency Accuracy & \bf Social Accuracy\\ \hline
Raw Super Characters method & 85.30\% & 82.50\%\\ \hline
Raw Super Characters method change line & 85.70\% & 83.30\%\\ \hline
SEW-text-only & 85.90\% & 83.30\%\\ \hline
SEW-text-only-Attention-Four-words & 86.00\% & 83.30\%\\ \hline
SEW-text-only-and-Profile-Features & 86.30\% & 85.60\%  \\ \hline
SEW-text-only-Attention-Profile-Features & \bf{86.90\%} & \bf{85.80\%}  \\ \hline
\end{tabular}
\end{center}
\caption{\label{Table_2} Results of raw Super Characters method and Squared English Word (SEW) method on HappyDB data set. For SEW method, attention scheme and profile information are also added. The model used are SE-net-154\cite{SENet154}. Text only means use only happy moment text information. User profile information includes age, country, marriage, and gender.}
\end{table*}
Compared to the original Super Characters method, SEW encapsulates one English word per square, rather than one English letter per square. The first word in the sentence goes in the top left square, and the succeeding words follow sequentially from left to right, proceeding onto the next row if necessary. Any remaining space is left empty, as a blank square. 

In Figure \ref{EnglishTextDemonstration_c}, the input image consisted of 6x6 squares, and the SEW Super Characters image is generated by only utilizing the happy moment text information. To distinguish from the other approaches below, we call this the SEW-text-only approach. 

In Figure \ref{EnglishTextDemonstration_d}, we also introduced an attention-based approach to make our model focus on particular important words or phrases within the input, such as, the first four words of the sentence. By allocating larger sized squares for the Super Characters that would hold certain English words, the convolutional layers within our model naturally dedicate greater emphasis on such words. This is common in the real world when we see signs and emphasized portion is enlarged to take attention as seen in Figure \ref{Fig_Street}. Similarly, people pay more attention to headlines than regular text in newspapers. 

We call the approach in Figure \ref{EnglishTextDemonstration_d} as SEW-text-only-Attention-Four-words, which applies the attention-based mechanism with an 8x8 input image with text only information in the happy moment. We chose to teach the network to pay particular attention to the first four words of a sentence, to see if the first four words have a large impact on the overall meaning of the sentence. With this specific implementation of the attention mechanism, we made the first four words two times the size of the rest of the words in the sentence, and positioned it on the center of the image. The regularly sized sentence flows as before, starting at the leftmost square of a row, continuing rightward on all possible places that can contain a squared English word until it hits the rightmost side of the row, then proceeding onto following rows. 

In Figure \ref{EnglishTextDemonstration_f},  we also use the profile features and happy moment text together. We set the profile features in Figure \ref{EnglishTextDemonstration_f} as the same size as the happy moment text information. Therefore, the resulting image is a combination of raw text input of happy moment and user-provided profile information. We call this approach as SEW-text-only-and-Profile-Features.

In Figure \ref{EnglishTextDemonstration_e}, similarly, we use both the user profile information and happy moment into the Super Characters image. And we also utilize attention scheme for the user profile information. We call this approach as SEW-text-only-Attention-Profile-Features. By using XGBoost \cite{chen2016xgboost} variable importance analysis tool, the parenthood information was determined to be the least important feature in classifying either social or agency when using only the profile information. So we only use four features from profile information, which are age, country, marriage, and gender. For age and country, we use the value as a single word. For marriage, we use initials of category values as the character to draw in the Super Character image, i.e. m (married), d (divorce), s (single), p (separated), w (widow), 0 ("nan"), and leave it empty for empty items. Similar for gender, f (female), m (male), o (other), and N ("nan").

\section{Experiments}
\subsection{CL-AFF Shared Task One}
We focused on the above mentioned six approaches as illustrated in Figure \ref{EnglishTextDemonstration} for training 2D-CNN models that could discern the agency and social tags of a given happy moment. 

For each approach detailed, we trained models by labeling the images with respect to social and agency values. Two separate datasets were created for the training of two different models. 

We randomly split the given labeled data into train and test at a ratio of 80\%:20\%. The histogram of word length distribution is given in Figure \ref{Fig_TrainData_histogram} for CL-AFF Train dataset, and  Figure \ref{Fig_TestData_histogram} for CL-AFF Test dataset.

The statistics of training and testing data set are given in Table \ref{Table_stats}.

Based on the statistics above, we set the cutlength at 36 for SEW-text-only as in Figure \ref{EnglishTextDemonstration_c}. For the 1.39\% of the sentences in the labeled data that contained more than 36 words, the 37th word and onwards were not included in the input image. In the shared task 170k test data set, 1.91\% were not included. 

For the attention method, we predefine an 8x8 two-dimensional array to act as the blueprint for the image inputs. There are zeros on all locations that are not designated for the special attended words, to indicate the positions allocated for such words. Of the space reserved for the attended words, all the values are -1 except for the top left box, which is of value 1.
Then, we will iterate through every square on the input image. If the corresponding value on the blueprint array based on the given indices a 0, we will draw the next English word in the input sentence using the SEW method. Should the value be 1, we will draw the SEW words in a larger font, and if it is -1, we will skip this iteration of the loop.

Table \ref{Table_2} shows our result based on a split of labeled data into 80\%:20\% for training and validation. The 2D-CNN model used are all SE-Net-154 \cite{SENet154}.

Comparing SEW-text-only with the raw Super Characters method with line change, we see a little accuracy improvement on agency prediction, from 85.7\% to 85.9\%. Social accuracy improved from 82.5\% to 83.3\% compared with raw Super Characters method without line change, although there is no improvement if comparing SEW-text-only to the raw Super Characters with line change. Although we did not see significant accuracy improvement by using SEW method in this data set, it did help improve accuracy by 2.1\% for the Wikipedia dataset as shown in Table \ref{Table_1}. The main reason for no significant accuracy improvement on this CL-Aff shared task data, is because the data size is not big enough. The CL-Aff data only has a total of 10,560 training samples for different categories, whereas the Wikipedia data set has 560,000 samples for training. Since the generated SEW Super Characters images are fed into CNN models to train, significant accuracy improvement will be observed for large data set because larger data sets help train better CNN models.

For SEW-text-only and SEW-text-only-Attention-Four-words, we see 0.1\% accuracy gain on agency label prediction by using attention in this data set, and we see no improvement for social prediction. Using other words to focus instead of using only the first four words, may further improve the accuracy. For example, we can use third party tools to extract keywords related to social or agency, then emphasize these words by enlarging them in the SEW image. Also, for a person's profile information, like age, country, marriage, and gender, the approach of SEW-text-only-Attention-Profile-Features embed them in the attention area, e.g. put the gender, marriage and etc. information in the attention area. 

The significant accuracy improvement for social prediction occurs when we add profile features into the SEW Super Characters image, which jumps from 83.3\% to 85.6\%. And the agency prediction accuracy also improves from 86.00\% to 86.3\%. After we further put these profile features into attention, it improves accuracy for both Agency and Social predictions. SEW-text-only-Attention-Profile-Features approach gives the best accuracy of 86.9\% for agency prediction, and also the best accuracy of 85.8\% for social prediction.

\section{Conclusion}
This paper borrows several ideas from Super Characters and attention, and we created a squared glyph for each English word. This Squared English Word (SEW) method can be trivially applied to other Latin languages. The structured user profile information and unstructured natural language information are integrated during preprocessing before sending CNN model to predict. 

\bibliographystyle{ACM-Reference-Format}
\bibliography{sample-base}


\end{document}